\documentclass[conference]{IEEEtran}
\IEEEoverridecommandlockouts
\usepackage{cite}
\usepackage{amsmath,amssymb,amsfonts}
\usepackage{algorithmic}
\usepackage{graphicx}
\usepackage{textcomp}
\usepackage{xcolor}
\usepackage{colortbl} 
\usepackage{multirow}
\usepackage{balance}

\def\BibTeX{{\rm B\kern-.05em{\sc i\kern-.025em b}\kern-.08em
    T\kern-.1667em\lower.7ex\hbox{E}\kern-.125emX}}
\begin{document}

\title{Towards Optimal Trade-offs in Knowledge Distillation for CNNs and Vision Transformers at the Edge
\thanks{This work was funded by AI4Media - A European Excellence Centre for Media, Society, and Democracy (EC, H2020 n.951911) and ELIAS - European Lighthouse of AI for Sustainability (EC, HE n.101120237}
}

\author{\IEEEauthorblockN{John Violos}
\IEEEauthorblockA{\textit{Information Technology Institute } \\
\textit{CERTH}\\ 
Thessaloniki, Greece \\
violos@iti.gr}
\and
\IEEEauthorblockN{Symeon Papadopoulos}
\IEEEauthorblockA{\textit{Information Technology Institute} \\
\textit{CERTH}\\ 
Thessaloniki, Greece \\
papadop@iti.gr }
\and
\IEEEauthorblockN{Ioannis Kompatsiaris}
\IEEEauthorblockA{\textit{Information Technology Institute} \\
\textit{CERTH}\\ 
Thessaloniki, Greece \\
ikom@iti.gr}
}

\maketitle

\begin{abstract}
This paper discusses four facets of the Knowledge Distillation (KD) process for Convolutional Neural Networks (CNNs) and Vision Transformer (ViT) architectures, particularly when executed on edge devices with constrained processing capabilities. 
First, we conduct a comparative analysis of the KD process between CNNs and ViT architectures, aiming to elucidate the feasibility and efficacy of employing different architectural configurations for the teacher and student, while assessing their performance and efficiency.
Second, we explore the impact of varying the size of the student model on accuracy and inference speed, while maintaining a constant KD duration.
Third, we examine the effects of employing higher resolution images on the accuracy, memory footprint and computational workload. 
Last, we examine the performance improvements obtained by fine-tuning the student model after KD to specific downstream tasks. 
Through empirical evaluations and analyses, this research provides AI practitioners with insights into optimal strategies for maximizing the effectiveness of the KD process on edge devices.
\end{abstract}

\begin{IEEEkeywords}
Knowledge Distillation, Vision Transformers, Frugal Learning, Fine Tuning
\end{IEEEkeywords}

\section{Introduction}
\label{Sec:Intr}
The proliferation of IoT and smart devices has sparked a continuous generation of visual data, revolutionizing how we interact with technology, take decisions and live in smart environments. From commodity cameras to specialised vision sensors, these devices capture a wealth of visual information, ranging from environmental observations to daily activities. 
By harnessing deep learning algorithms on the edge, we can process and analyze visual data in real time, enabling new capabilities for end users. Keeping data on the edge offers advantages, particularly regarding privacy, security, and data transmission, but it also imposes computational constraints. 

CNNs and ViTs are the cornerstone architectures for various computer vision tasks, exhibiting state-of-the-art performance on numerous benchmarks. Nevertheless, their computational demands often render them impractical to train or even for deployment on resource-constrained edge devices. On the other hand, there is currently wide availability of powerful pre-trained deep learning vision models, which have been trained extensively on vast datasets. Such models are often used as \textit{teachers} for transferring knowledge to smaller, compressed \textit{student} models through \textit{knowledge distillation}. 
Efficient knowledge distillation from teachers to students can facilitate and streamline the development of computationally frugal computer vision systems.

In this way, KD has emerged as a promising approach for model compression, wherein a compact student Artificial Neural Network (ANN) is trained to mimic the behavior of a larger, more complex teacher ANN. While this technique has been widely applied using a large amount of computational resources provided by cloud infrastructures and data centers, its application to limited resources and its effectiveness in edge computing settings, especially for visual data, remains relatively unexplored. 
It is noteworthy that our research focuses on both optimizing the KD process for lightweight implementation and ensuring the accuracy and efficiency of the resulting student model.
From the early steps of our research we noted that student models derived from established KD processes demonstrate efficient execution capabilities on the edge. However, the KD process was also found to be computationally intensive rendering its feasibility questionable in such resource-constrained environments.
This paper endeavors to fill the research gap regarding the applicability of KD to ANN vision models deployed on computationally-constrained resources.

In our work, edge computing refers to the paradigm of decentralized data processing where training, inference and even the compression of ANNs are performed closer to the data capturing end-user devices. 
The edge computing environments we consider are confined spaces, such as smart homes, wherein computational tasks are executed utilizing one single commercially Graphics Processing Unit (GPU).
We discuss and conduct experiments on benchmark datasets along the following four facets of KD: 

\begin{enumerate}
\item The comparison of the KD process between CNNs and ViT architectures. 
\item The exploration on how the size of the student model affects the accuracy, inference and KD time. 
\item The exploration of the improvement in accuracy and increase in inference and KD time when we use higher resolution images as input. 
\item The investigation of performance improvement  after fine-tuning the student model. 
\end{enumerate}

Section \ref{Sec:RW} discusses the related work and highlights existing research gaps. 
In Section \ref{Sec:PropMod}, we present our methodology.
Section \ref{Sec:ExpEval} describes our experiments and results. Finally, Section \ref{Sec:Conc} concludes the paper and provides some future directions.

\section{Related Work \& Background}
\label{Sec:RW}
In the pursuit for enhanced performance, state-of-the-art CNNs architectures have progressively evolved towards greater complexity, thereby necessitating substantial memory resources and computational power. Consequently, the deployment of such models on the edge, i.e. limited resource settings, poses considerable challenges \cite{li_model_2023}. In this direction Chen et al. \cite{chen_review_2023} present and review the main approaches for the design of lightweight CNNs that include: i) architectural design, ii) model compression, iii) general learning libraries, and iv) hardware support. 

The scope of our research belongs to the field of model compression. 
Model compression include pruning, low-rank decomposition, low-bit quantization and KD. Cheng et al. \cite{cheng_survey_2020} reviewed compression techniques and concluded that KD methods can be advantageous compared to the alternatives in cases where small/medium datasets are used or where  significant improvements in terms of efficiency are sought.  
Through KD, the compressed student model can leverage the  transferred knowledge from the teacher, attaining competitive performance, especially in cases where small/medium-sized datasets are available. This prompted us to focus our research on KD methods.

Lin Wang and Kuk-Jin Yoon \cite{wang_knowledge_2022}  provided a comprehensive overview of literature concerning the utilization of KD in computer vision tasks. They outlined various taxonomies of different KD methods and elucidated their application across various computer vision scenarios. Moreover, they delved into fundamental questions such as the optimal size of the teacher model, the significance of using a pre-trained teacher, the potential enhancements achieved through an ensemble of teachers, and the efficacy of data-free distillation techniques. In the end, they posed an unresolved research question concerning the potential impact of varying ANN architectures between teacher and student on the efficacy of KD. This is connected with the first facet in our study.

Large ANN models with a greater number of parameters tend to exhibit high accuracy due to their increased capacity to capture complex patterns and relationships in the data. However, this often comes at the cost of longer inference times, as larger models require more computational resources. Conversely, small models with fewer parameters may offer reduced accuracy compared to their larger counterparts but typically entail faster inference times, making them more suitable for edge computing settings. Thus, striking a balance between model size, accuracy, and inference time is paramount when designing ANNs. While there are studies \cite{li_train_2020} that extensively discuss the impact of model size and the trade off between the accuracy and the learning time for ANN pruning, to the best of our knowledge we are the first who conduct this research in the context of KD as part of the second facet of our analysis.

Habib et. al \cite{habib_knowledge_2023} conducted a comprehensive review focusing on KD in ViT. Specifically, they examined 30 KD approaches, categorized into nine primary categories, with some of the most prominent being the Target-aware Transformer, the Fine-Grain Manifold Distillation Method, the Cross Inductive Bias Distillation, Tiny-ViT, and the Attention Probe-based Distillation Method. The authors conclude that ViTs exhibit promising results comparable to or even surpassing the state-of-the-art CNN architectures, but they entail high training and inference costs, rendering them unsuitable for resource-constrained devices. Moreover, the KD process itself incurs significant training and inference costs. A research gap that the authors identify concerns the impact of image resolution on the KD process, which is addressed in the third facet of our study.

Fine-tuning stands as a widely embraced methodology employed in the framework of ANN pruning and transfer learning  \cite{han_dsd_2017}. Through fine-tuning, pruned ANNs or pre-trained models undergo further training on downstream task-oriented datasets under the assumption that those are available in an edge environment. In this way, large generic models are specialized to better suit the intricacies of the target task data. Despite its prevalent use, an aspect that remains relatively unexplored is the application of fine-tuning in the student model derived by the KD process. The exploration of fine-tuning student models when the KD process runs in limited computational and data settings is addressed by the fourth facet of our study.

\section{Four facets for KD on the Edge}
\label{Sec:PropMod}

\subsection{Different ANN Types for Student and Teacher}
KD involves transferring knowledge from a teacher ANN model to a student ANN model by minimizing the discrepancy between the logits produced by each model. Unlike traditional compression methods, KD can effectively reduce the size of an ANN regardless of structural and typological differences between the teacher and student architecture. The structural dissimilarity between the teacher and student architectures significantly influences the student's performance and the efficacy of the distillation process. Opting for a complex ANN architecture, such as a ViT, the teacher model is anticipated to yield greater robustness. This happens because the knowledge (soft labels) imparted by an advanced teacher can more accurately capture class distributions. However, the use of complex ANNs also incurs longer inference times, thereby impeding the timely flow of information during the distillation process and elevating computational costs.

Similar considerations apply to the selection of the student model. While prior research \cite{habib_knowledge_2023} suggests that ViTs outperform CNNs, one should note that transformers necessitate substantially more time and computational resources for training. While this may not pose a significant challenge in training scenarios supported by powerful cloud infrastructures, it becomes a critical concern in more constrained environments. Given this consideration, the choice of ANN architecture for student and teacher in order for the student to deliver accurate outcomes with limited distillation time and resources warrants further experimental investigation.

\begin{table}[]
\centering
\caption{Knowledge Distillation between ViTs \& CNNs}
\begin{tabular}{|l|c|c|c|}
\hline
{\color[HTML]{000000} Teacher} & {\color[HTML]{000000} Student} & {\color[HTML]{000000} Acc.} & {\color[HTML]{000000} Learning (KD) Time} \\ \hline
CNN                            & CNN                            & Best                        & Best                                      \\ \hline
CNN                            & Transformer                    & Bad                         & Good                                      \\ \hline
Transformer                    & CNN                            & Good                        & Bad                                       \\ \hline
Transformer                    & Transformer                    & Worst                       & Worst                                     \\ \hline
\end{tabular}

\label{tab_diff_arch}
\end{table}

\subsection{Student Model Capacity}
The gap of model capacity between the teacher and the student is a critical consideration in the KD process. An optimal balance must be struck to ensure effective distillation outcomes. When the student model's capacity is too low relative to the teacher, the student struggles to effectively incorporate the logits information provided by the teacher. This limitation impedes the student's ability to capture the distilled knowledge, thus hindering its performance. 

Conversely, when the student model's capacity is excessively large, the expected improvements in distillation efficacy may not materialize. Larger student models tend to exhibit slower learning rates and are more susceptible to overfitting, resulting in diminished generalization performance. Furthermore, the computational and memory resources required for training and deploying larger student models pose significant challenges, particularly in resource-constrained environments such as edge devices. The increased computational demands, higher memory footprint, and slower inference speeds associated with larger models render them less practical for deployment in real-world edge scenarios. Therefore, determining the optimal student size also requires experimental investigation.

\subsection{Higher Resolution Images}
The resolution of input images significantly influences the KD process, impacting both model performance and computational efficiency. When utilizing low-resolution images, computational demands are reduced, leading to expedited distillation and inference times. 
However, this comes at the cost of potential loss of fine-grained details and contextual information, which may hinder model performance, particularly in tasks requiring precise object recognition or classification. Furthermore, low-resolution images may limit the model's ability to generalize to diverse data distributions and discriminate between similar classes.

Models trained on high-resolution images exhibit improved generalization capabilities and enhanced discriminative power, leading to more robust performance. Nonetheless, the computational complexity associated with processing high-resolution images results in longer training and inference times, posing challenges in resource-constrained environments. Additionally, high-resolution images may be more sensitive to noise. This affects model training and performance, while also requiring higher memory resources during deployment.

\subsection{Fine-tuning the Student Model}
Fine-tuning is a widely established technique that is very often employed in compressed models with the aim of enhancing their performance. However, it is important to note that fine-tuning also entails significant time and computational resource consumption. Often likened to a second training stage, fine-tuning necessitates careful consideration from AI practitioners regarding the trade-offs involved. The computational overhead incurred by fine-tuning prompts a reevaluation of its utility and efficacy in optimizing model performance, particularly in resource-constrained environments.

Furthermore, the significance of fine-tuning is heightened in edge computing scenarios. Edge devices, situated at the periphery of the network, have the capability to capture  images that are more relevant to the specific user context. Leveraging fine-tuning with these locally captured images enables the adaptation of ANNs to specific needs. This tailored approach not only enhances model performance but also justifies the additional computational resources required for fine-tuning. 

\section{Experimental Evaluation}
\label{Sec:ExpEval}
We conducted experiments utilizing various transformer architectures, including the basic ViT and more advanced architectures such as Data-efficient Image Transformer (DeiT), and Swin Transformer \cite{huynh_vision_2022}, alongside the classic VGG \cite{simonyan_very_2015} CNN architecture. 
We experimented with various combinations of these models, ensuring that the teachers always had significantly more parameters than the students. For student models, in addition to the standard VGG variations with 11, 13, 16, and 19 layers described in \cite{simonyan_very_2015}, we also conducted experiments with models having 2, 3, 4, 5, and 8 layers.

\begin{table*}[]
\centering
\caption{Exploring Knowledge Distillation with Various Sizes of CNN Students \& Image Resolutions}
\begin{tabular}{|l|l|l|l|l|l|l|}
\hline
                               &           & Extra Small           & Small      & Medium      & Large      & Extra Large                \\ \hline
                               & Params (M) & 0.004178          & 0.015978     & 0.062474     & 0.24705      & 0.982538              \\ \hline
\multirow{2}{*}{Img\_size:32}  & Accuracy  & 0.591099      & 0.730499 & 0.824799 & 0.875299 & \textbf{0.906099} \\ \cline{2-7} 
                               & Ops (M)    & \textbf{0.163754} & 0.511818     & 1.760906     & 6.470922     & 24.738314             \\ \hline
\multirow{2}{*}{Img\_size:224} & Accuracy  & 0.668199      & 0.767099 & 0.824999 & 0.861899  & 0.878799          \\ \cline{2-7} 
                               & Ops (M)    & 8.015786          & 25.063242    & 86.253194    & 317.013258   & 1212.054026           \\ \hline
\end{tabular}

\label{tab_students_sizes}
\end{table*}

We opted to utilize the VGG architecture over other state of the art architectures such as EfficientNet and ResNet due to its flexibility in accommodating various sizes, allowing us to investigate the impact of different student sizes in our research.
These experiments were carried out using three popular datasets: Cifar10, Cifar100, and the large-scale ImageNet-1k dataset. 
We utilized the KD approach implemented in the Model Compression Research Package \cite{zafrir_ofir_2021_5721732} by Intel Labs, which is based on the original paper by Geoffrey Hinton \cite{hinton_distilling_2015}.
Utilizing a consumer GPU equipped with 8 GB of memory, we systematically evaluated the performance and efficiency of each model architecture across these datasets. By employing a range of datasets, transformers and VGG, our experiments aim to  shed light on the four facets of KD for computer vision tasks as described in previous sections.

Before embarking on experiments involving KD, we conducted investigations into training CNN and transformer architectures from scratch on a GPU for limited time periods that range from few up to 24 hours.
Our observations revealed that training a CNN model from scratch on a GPU yielded superior performance compared to training a transformer. For the same training periods the transformers' accuracy was lower compared to the one attained by CNN from 4\% to 16\%. 
Despite employing the same dataset across all experiments, the transformer's training process exhibited a substantially slower learning rate, proving inefficient when constrained to the limited computational setting of a single GPU. 


\subsection{Experiments on Different ANN Types for Student-Teacher}

Through experiments with ViT, DeiT, Swin and VGG on Cifar10, Cifar100, and ImageNet-1k, we aimed to explore the efficacy and intricacies of distilling knowledge between established CNNs models and transformer architectures. 
Specifically, we explored the effectiveness of KD: i) from CNNs to CNNs, ii) from CNNs to Transformers, iii) from Transformers to CNNs, and iv) from Transformers to Transformers examining how different models for teachers and students impact the performance (Acc) and efficiency (KD Time) of the distillation process and the compressed models. 
A qualitative summary of the obtained results is presented in Table \ref{tab_diff_arch}.

The experimental outcomes underscore that the KD process exhibits differing dynamics when applied between different models for the teacher and student. We observed that the KD process is faster and leads to the best accuracy when implemented between CNNs due to the swift output of logits and the distillation loss calculations. 
Conversely, transformers require numerous epochs and iterations to effectively learn during the distillation process. This disparity arises primarily from the intricate architecture and larger capacity of transformers compared to CNNs. 
The transformer's higher capacity result in longer inference times and make slower the KD processes compared to CNNs. 
Consequently, the KD process with transformers is characterized by prolonged learning periods and slower inference times, contrasting with the swifter and more efficient KD process observed with CNNs.

In further experiments on Transformers, we also allowed the distillation to run for several days. However, the accuracy improvements were very slow to the point that we considered them impractical when considering an edge environment as a target deployment setting for the KD process. 

\subsection{Experiments on the Student Model Capacity}
This subsection delves into the experimental investigation of how varying the size of the student CNN model influences the accuracy and inference time. 
Through experimentation with student models of different capacities, ranging from compact to more expansive architectures, we aimed to elucidate the trade-offs between model size and performance. Table \ref{tab_students_sizes} presents a set of representative experimental outcomes when performing KD from DeiT to VGG. The different sizes of VGG are estimated by the million (M) of their parameters (Params). Moreover, we examine the implications of model size on the efficiency of the KD process, including the number of operations (Ops) for student inference.

Through quantitative analysis of accuracy metrics in conjunction with the capacity of student models, it becomes evident that larger CNN models for students, characterized by a higher number of parameters, tend to exhibit superior accuracy. However, this is accompanied by a corresponding increase in computational complexity, as larger models entail a greater number of operations. Specifically, the Extra Large student model comprising approximately 98 million parameters, necessitates approximately 24 million operations for inference while achieving 90\% accuracy. 
The observations in this and previous subsections highlight the value of allocating edge resources for a relatively large CNN model as a student, in contrast to the use of transformers.

\subsection{Experiments on Higher Resolution Images}
We investigated the impact of employing higher resolution images on the accuracy, workload and memory resources. In Table \ref{tab_students_sizes}, we present a series of experiments conducted with image resolutions 32 x 32 and 224 x 224. 
In both cases we kept the duration of KD constant. The analysis of results reveals an interesting trend: For student models with smaller capacities (extra small and small), employing higher resolution images leads to improved accuracy. However, this pattern is reversed for larger student models (large and extra large), where lower resolution images yield higher accuracy. This phenomenon can be attributed to the slower learning rates observed in larger student models when processing higher resolution images. The intricacies of high-resolution data introduce greater complexity to the learning process, causing larger models to struggle to converge effectively and ultimately resulting in lower accuracy compared to their smaller counterparts.

In terms of memory footprint, a single RGB image with a resolution of 32 x 32 pixels and 24 bits per pixel (bpp) consumes 3072 bytes, while the same image at a resolution of 224 x 224 pixels requires 150,528 bytes. Similarly, grayscale images with 8 bpp exhibit notable differences in size, with a 32 x 32 resolution image occupying 1024 bytes and a 224 x 224 resolution image consuming 50,176 bytes. These numbers underscore that the memory requirements for both RGB and grayscale images are substantially larger, approximately 49 times, for the 224 x 224 resolution compared to the 32 x 32 resolution. A similar trend is observed in computational workload, where operations required for 224 x 224 images range from 49 to 52 times those required for their 32 x 32 counterparts. 
These findings lead us to conclude that utilizing relatively large CNN models for students while employing lower image resolutions achieves the optimal balance between performance, memory usage, and computational efficiency.

\subsection{Experiments with Fine-tuning the Student Model}

\begin{table}[]
\centering 
\caption{Enhancement in Accuracy with Fine-Tuning on CNNs \& ViTs}
\begin{tabular}{|l|l|cc|}
\hline
       &            & \multicolumn{2}{c|}{Accuracy}                                                  \\ \hline
       & Params (M) & \multicolumn{1}{l|}{Prior Fine-Tuning} & \multicolumn{1}{l|}{Post Fine-Tuning} \\ 
       \hline
VGG & 24.73       & \multicolumn{1}{c|}{0.9060}            & 0.9342                                \\
       \hline
Swin-T & 26.60       & \multicolumn{1}{c|}{0.8475}            & 0.9616                                \\ \hline
ViT    & 85.27      & \multicolumn{1}{c|}{0.8600}            & \textbf{0.9814}                                \\ \hline
\end{tabular}
\label{tab_fine_tuing}
\end{table}

Fine-tuning in ViT and VGG-CNNs involves selectively training certain parts of the student model while keeping other parts frozen to preserve the knowledge captured during KD. In ViT the process starts by freezing the transformer layers to maintain the learned representations and only fine-tuning the task-specific layers namely the final classification head. Similarly, in VGG-CNNs, the convolutional layers are kept frozen to retain their  feature extraction capabilities, while the fully connected layers and fine-tuned. The learning rate, which controls the step size at each training iteration, is set to lower values during fine-tuning to ensure that the adjustments to the pre-trained weights are limited and do not disrupt existing knowledge gained by KD.

We present our findings on the impact of fine-tuning the student model derived from the KD process in Table \ref{tab_fine_tuing}. It is evident that fine-tuning transformers results in significant improvements compared with CNNs. Specifically fine-tuning the Swin-T model led to a substantial increase in accuracy of approximately 16\%. Similarly, when fine-tuning the ViT model, we observed a notable improvement of almost  12\% in accuracy. 
On the other hand, we observed only small improvements (approximately 3\%) when fine-tuning the VGG model. 
These were consistently observed across all models when utilizing the CIFAR-10 dataset. We used different parts of the dataset for distillation, fine-tuning and evaluation. In all experiments, the fine-tuning process occurred over the course of 15 epochs.

\section{Conclusions and Future Work}
\label{Sec:Conc}
In this research paper, we offer insights for researchers and practitioners aiming to compress and deploy ANNs in edge environments under constrained computational resources and time limitations. Despite the prevailing perception of Transformers' superior accuracy, our empirical evidence indicates that utilizing relatively large CNNs with low resolution images presents a more efficient and feasible approach to derive highly performing and efficient compressed student models. Only when fine-tuning is applied after KD, we observe that transformer students surpass CNNs, albeit with a significant increase in resource consumption and time delay.
As future work our aim is to design a lightweight KD methodology that takes into consideration the four facets outlined in this paper and addresses the computational bottlenecks using a meta-heuristic approach. This will be based on a smart search of the hypothesis space of potential student architectures, aiming to identify the closest-to-optimal solution based on criteria such as KD time, student inference time, resource utilization, and energy consumption.

\section*{Acknowledgment}
This work was funded by AI4Media - A European Excellence Centre for Media, Society, and Democracy (EC, H2020 n.951911) and ELIAS - European Lighthouse of AI for Sustainability (EC, HE n.101120237) 

\balance
\bibliographystyle{IEEEtran}
\bibliography{References.bib}

\end{document}